\documentclass[letterpaper]{article} % DO NOT CHANGE THIS
\usepackage{aaai2026}  % DO NOT CHANGE THIS
\usepackage{times}  % DO NOT CHANGE THIS
\usepackage{helvet}  % DO NOT CHANGE THIS
\usepackage{courier}  % DO NOT CHANGE THIS
\usepackage[hyphens]{url}  % DO NOT CHANGE THIS
\usepackage{graphicx} % DO NOT CHANGE THIS
\usepackage{natbib}  % DO NOT CHANGE THIS AND DO NOT ADD ANY OPTIONS TO IT
\usepackage{caption} % DO NOT CHANGE THIS AND DO NOT ADD ANY OPTIONS TO IT
\usepackage{algorithm}
\usepackage{algorithmic}
\nocopyright
\usepackage{newfloat}
\usepackage{listings}
\usepackage{amsfonts}
\usepackage{tabularx}
\usepackage{amsmath}
\usepackage{booktabs}
\usepackage{dsfont}
\usepackage{multirow}
\usepackage{makecell}
\usepackage{bm}

\DeclareCaptionStyle{ruled}{labelfont=normalfont,labelsep=colon,strut=off} % DO NOT CHANGE THIS
\floatstyle{ruled}
\newfloat{listing}{tb}{lst}{}
\floatname{listing}{Listing}
\title{Flow-Aware GNN for Transmission Network Reconfiguration via Substation Breaker Optimization}
\author {
    Dekang Meng\textsuperscript{\rm 1},
    Rabab Haider\textsuperscript{\rm 2},
    Pascal van Hentenryck\textsuperscript{\rm 1}
}
\affiliations {
    \textsuperscript{\rm 1}H. Milton Stewart School of Industrial and Systems Engineering, Georgia Institute of Technology\\
    \textsuperscript{\rm 2}Department of Civil \& Environmental Engineering, University of Michigan\\
    dm649@gatech.edu, rababh@umich.edu, pvh@gatech.edu
}
\begin{document}
\maketitle

\begin{abstract}
This paper introduces \textsc{OptiGridML}, a machine learning framework for discrete topology optimization in power grids. The task involves selecting substation breaker configurations that maximize cross-region power exports, a problem typically formulated as a mixed-integer program (MIP) that is NP-hard and computationally intractable for large networks. \textsc{OptiGridML} replaces repeated MIP solves with a two-stage neural architecture: a line-graph neural network (LGNN) that approximates DC power flows for a given network topology, and a heterogeneous GNN (HeteroGNN) that predicts breaker states under structural and physical constraints. A physics-informed consistency loss connects these components by enforcing Kirchhoff's law on predicted flows. Experiments on synthetic networks with up to 1{,}000 breakers show that \textsc{OptiGridML} achieves power export improvements of up to 18\% over baseline topologies, while reducing inference time from hours to milliseconds. These results demonstrate the potential of structured, flow-aware GNNs for accelerating combinatorial optimization in power grid systems.
\end{abstract}

\section{Introduction}

Transmission network reconfiguration (TNR) is an operational strategy in high-voltage power grids that alters the network topology by opening or closing switches and circuit breakers. Traditionally used for service restoration and contingency management, TNR has recently become critical for addressing transmission congestion, particularly to enable efficient power transfer from generation-rich regions to major load centers. Congestion not only limits renewable energy utilization, but also contributes to significant spatial price disparities in electricity markets.

Substation-level bus splitting, achieved by selectively opening breakers within a substation, allows fine-grained adjustments of network topology by changing bus connectivity. Unlike conventional line switching, which removes existing transmission lines, bus splitting structurally alters the grid's nodal graph, resulting in significant system-wide impacts on power flows. For instance, operators may seek to increase power transfers from wind-abundant regions (source zone) to high-demand urban areas (sink zone) by strategically reconfiguring substations to alleviate congestion on key transmission corridors. This capability to actively manage cross-zonal transfer is crucial for integrating variable renewable generation and improving market efficiency.

Currently, TNR decisions are made primarily manually by grid operators. However, the increasing spatiotemporal complexity of modern grids motivates the development of automated decision-support tools that quickly identify effective breaker configurations. Although optimal breaker configurations for bus splitting can be formulated as a mixed-integer program (MIP), solving such formulations is NP-hard and computationally infeasible for {\em large-scale, real-time operational decisions}. Prior research has explored machine learning (ML) methods for accelerating related optimization problems, including line switching and distribution system reconfiguration.

This paper proposes \textsc{OptiGridML}, a supervised learning framework designed specifically for predicting breaker configurations to maximize cross-zonal transfer capability. Instead of merely accelerating solver internals, \textsc{OptiGridML} serves as an \textit{optimization proxy} \cite{wenbo_UC_2023}, replacing repeated MIP solves with rapid inference via a learned model. The main contributions of this work are:
\begin{enumerate}
    \item \textsc{OptiGridML} introduces a two-stage graph neural network (GNN) framework, consisting of a line-graph neural network (LGNN) that approximates DC power flows and a heterogeneous GNN (HeteroGNN) that predicts breaker-level configurations. The line-graph representation ensures consistent input dimensionality under dynamic topology changes and directly models flow redistribution on transmission lines, a combination not widely explored in previous ML applications for power systems.
    
    \item A novel physics-informed consistency loss enforces Kirchhoff's Current Law (KCL) at busbars using approximate flows from the LGNN, even in the absence of explicit flow labels. Structural feasibility is further enhanced through constraint-aware auxiliary penalties addressing line overloads, invalid substation splits, and disconnected busbars, supplemented by a lightweight inference-time repair mechanism.
    
    \item The proposed framework achieves near-optimal breaker configurations on synthetic networks of up to 1,000 breakers, significantly improving power export capability (up to 18\%) compared to baseline configurations, while dramatically reducing inference times from over 10 hours to milliseconds.
\end{enumerate}

Together, these contributions demonstrate the important role ML can play in supporting grid operators seeking to advance quick and safe power transfers in large power grids.

\section{Related Work}

\paragraph{Traditional Approaches to TNR}

TNR has traditionally been formulated as a mixed-integer optimization problem involving line switching or substation-level breaker operations. Early efforts focus on topology control through line switching to reduce congestion and improve economic dispatch, using deterministic or security-constrained formulations \cite{ruiz2009effective, pandzic2014transmission, hedman2011review}. More recent efforts extend this framework to explicitly model breaker-level reconfiguration within substations, enabling bus splitting and finer-grained control of network topology \cite{fan2021transmission, sahoo2022breaker}. These methods solve for optimal configurations under DC or AC power flow constraints but face scalability challenges for large networks and repeated decision-making.

\paragraph{ML for Power Systems}

A significant body of research applies machine learning to power systems, primarily targeting optimal power flow (OPF) under linear DC or nonlinear AC models \cite{Zamzam_SmartGridComm2020, donti_DC3, deepopf, Fioretto_lagrangeduals, Qui_dualconicproxies_2024, piloto2024canosfastscalableneural, varbella2024physicsinformedgnnnonlinearconstrained}. However, most of these works assume fixed topology or react to topology changes without explicitly modeling them \cite{Chen_metalearning_topologyreconfig, piloto2024canosfastscalableneural} and focus on continuous decision spaces without addressing discrete configuration decisions central to TNR.
ML has also been explored for binary decision making in problems such as unit commitment (UC), where generator on/off scheduling is optimized \cite{Xavier_SCUC, deepopf_SCUC, wenbo_UC_2023, Bahrami_UC_2022}. While relevant in structure, many of these methods relax the discrete nature of the problem through approximations or surrogate objectives. In the context of distribution network reconfiguration (DNR), recent work applies supervised or reinforcement learning to accelerate search \cite{DyR_spacereduction_ISGT_2024, Qin_2023_GNNreconfig, DyR_spacereduction_TPWRS_2024}. A subset of these approaches incorporate discrete control decisions into the learning process directly, using physics-informed or constraint-aware models to preserve feasibility \cite{Haider_DNR_MLproxy, Authier_DNR_GNN}. More recently, Kim and Kim~\cite{kim2025dispatch} introduced a dispatch-aware deep neural network approach for optimal transmission switching, using differentiable DC-OPF layers to guarantee feasibility and achieve real-time inference. 

Although these advances offer promising directions, most prior work focused on tree-structured distribution networks, continuous control, or discrete line operations, leaving a gap in learning-based solutions for discrete substation reconfiguration in meshed transmission systems.

\paragraph{ML for Constraint Optimization}

The need for fast and repeated solutions to optimization problems has motivated a growing body of ML-based approaches. While earlier efforts focused on continuous programs, recent work increasingly targets MIPs. A major line of research aims to accelerate existing solvers through learned heuristics, such as neural diving \cite{Nair_MIPNN_2020} and neural branching policies \cite{Gasse_COGNN_2019, Khalil_MLbranching}. Reviews of ML for MIP include surveys of variable selection, cutting plane generation, and heuristics such as feasibility pumps \cite{Zhang_MIPML_survey}, as well as broader surveys on GNNs for combinatorial optimization \cite{Khalil_COGNN_2021}.

Beyond solver acceleration, recent methods aim to predict full or partial solutions to combinatorial problems. Some approaches generate high-confidence subsets of integer variables as a warm-start \cite{han2023a, pmlr-v235-huang24f}, while others learn full discrete solutions using differentiable correction layers \cite{Haider_DNR_MLproxy, tang2024learningoptimizemixedintegernonlinear}. Lu et al. \cite{lu2025boosting} propose a hybrid learning-and-optimization framework for trip and crew scheduling, where a GNN predicts promising subproblem columns that are then validated and refined by a classical solver. Inspired by this architecture, the proposed \textsc{OptiGridML} framework applies a similar principle in the context of TNR. A GNN-based predictor generates candidate substation-level breaker configurations, which are subsequently evaluated through DC-OPF solver to ensure physical and operational feasibility.

\section{Problem Statement}
\begin{figure}[tb]
    \centering
    \includegraphics[width=0.3\textwidth]{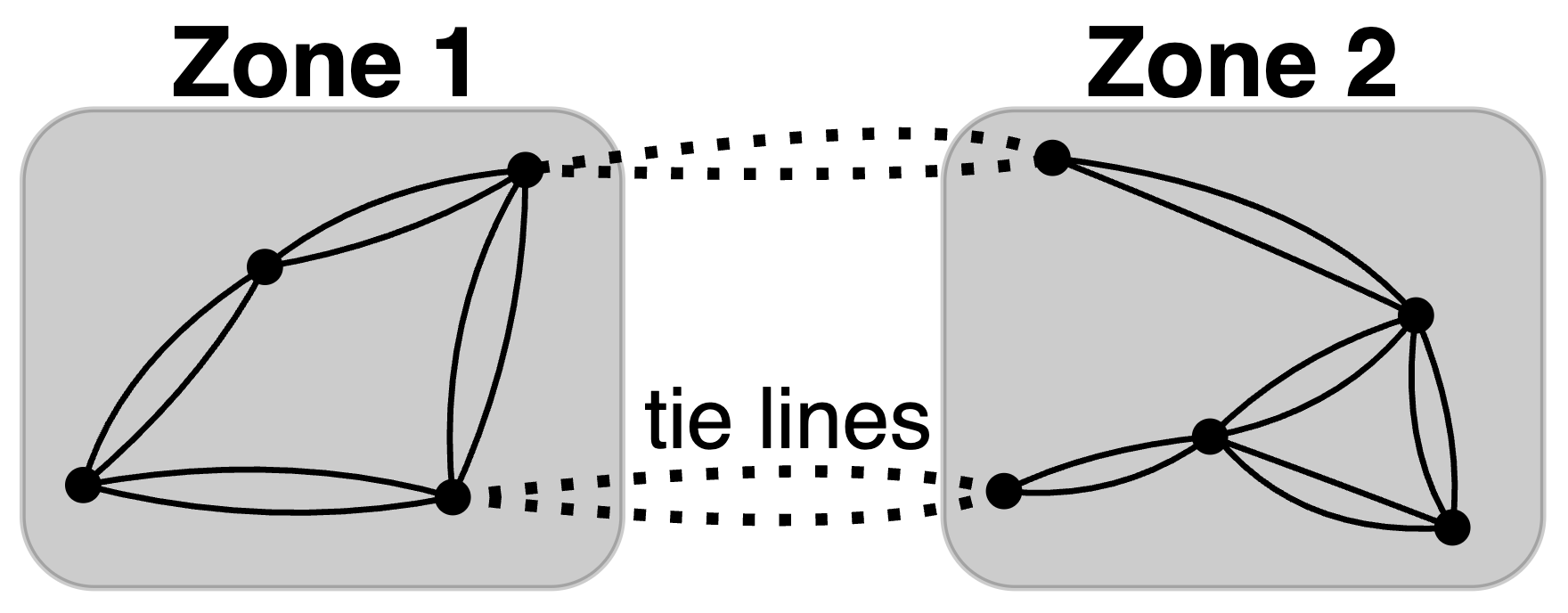}
    \caption{A $2$-zone power grid with double power lines.}
    \label{fig:zones}
\end{figure}

This work considers a two-zone transmission network, where the goal is to determine breaker configurations that maximize power exports from a generation-dominant source zone (Zone 1) to a load-heavy sink zone (Zone 2). Such two-zone architectures arise in scenarios involving energy islands, offshore grids, or regional interconnects, where bulk power must be transferred across limited interfaces. Representative examples include offshore wind farms or energy islands exporting electricity to onshore grids, a structure observed in the North Sea, Baltic Sea, and along the eastern seaboard of the United States. The two zones are connected via a small number of tie lines, while substations within each zone are connected by parallel transmission lines as illustrated in Figure~\ref{fig:zones}. Each substation contains an internal structure of busbars and breakers. Power injections are defined at the busbar level, with initial generation in Zone 1 and load in Zone 2 scaled proportionally to their base profiles. The objective is to reconfigure breaker states within substations to induce a new topology that supports greater export from Zone 1 to Zone 2, while satisfying power flow feasibility. The resulting topology must preserve system connectivity or maintain local power balance within any islanded components. A DC power flow model is assumed, in which flows on transmission lines are linearly related to voltage angles and subject to thermal capacity constraints. Breaker decisions directly affect the graph structure and determine feasible power flow pathways across the network. Consider a power grid with \( n \) substations and \( m \) transmission line pairs, represented by sets \( \mathcal{S} \) and \( \mathcal{L} \), respectively. The connectivity of the network is encoded in a matrix indicating which substations are connected by power lines. Table~\ref{tab:notation} summarizes the notation used in the formulation.

\begin{table}[!t]
\centering
\small
\begin{tabularx}{\columnwidth}{@{}lX@{}}
\toprule
\textbf{Symbol} & \textbf{Description} \\
\midrule
\multicolumn{2}{@{}l}{\textbf{Sets and Parameters}} \\
$\mathcal{B}$ & Set of busbars (indexed by $i$) \\
$\mathcal{L}$ & Set of transmission lines (indexed by $\ell$) \\
$\mathcal{R}$ & Set of breakers (indexed by $r$) \\
$\mathcal{S}$ & Set of substations (indexed by $s$) \\
$\mathcal{B}_s$, $\mathcal{R}_s$ & Busbars and breakers in substation $s$ \\
$\mathcal{B}_1$, $\mathcal{B}_2$ & Busbars in zones 1 and 2 \\
$\bar{P}_i$ & Base generation capacity at busbar $i$ \\
$d_i$ & Base load at busbar $i$ \\
$X_\ell$ & Reactance of line $\ell$ \\
$\bar{F}_\ell$ & Flow limit on line $\ell$ \\
$M$ & Big-M constant \\
\addlinespace[0.5em]
\multicolumn{2}{@{}l}{\textbf{Constants}} \\
$\alpha$ & Initial Zone 1 to Zone 2 generation-to-load ratio:
\\ & $\displaystyle \alpha = \frac{\sum_{i \in \mathcal{B}_1} \bar{P}_i}{\sum_{i \in \mathcal{B}_2} d_i}$ \\[0.3em]
$\beta$ & Normalized offset parameter for power balance:
\\ & $\displaystyle \beta = \frac{\sum_{i \in \mathcal{B}_2} \bar{P}_i - \sum_{i \in \mathcal{B}_1} d_i}{\sum_{i \in \mathcal{B}_2} d_i}$ \\
\addlinespace[0.5em]
\multicolumn{2}{@{}l}{\textbf{Decision Variables}} \\
$\lambda \in [0,1]$ & Generation capacity utilization ratio in zone 1 \\
$\mu \in [0,1]$ & Load capacity utilization ratio in zone 2 \\
$P_i^G \in [0, \bar{P}_i]$ & Generation at busbar $i$ \\
$P_i^D$ & Demand at busbar $i$ \\
$\theta_i \in [-\pi, \pi]$ & Voltage angle at busbar $i$ \\
$f_\ell \in [-\bar{F}_\ell, \bar{F}_\ell]$ & Power flow on line $\ell$ \\
$f_r \in \mathbb{R}$ & Flow across breaker $r$ \\
$z_r \in \{0,1\}$ & Breaker status (1 = closed, 0 = open) \\
\bottomrule
\end{tabularx}
\caption{Notation used throughout this TNR study.}
\label{tab:notation}
\end{table}

\paragraph{Power Grid Configuration}

\begin{figure}[tb]
    \centering
    \includegraphics[width=0.45\textwidth]{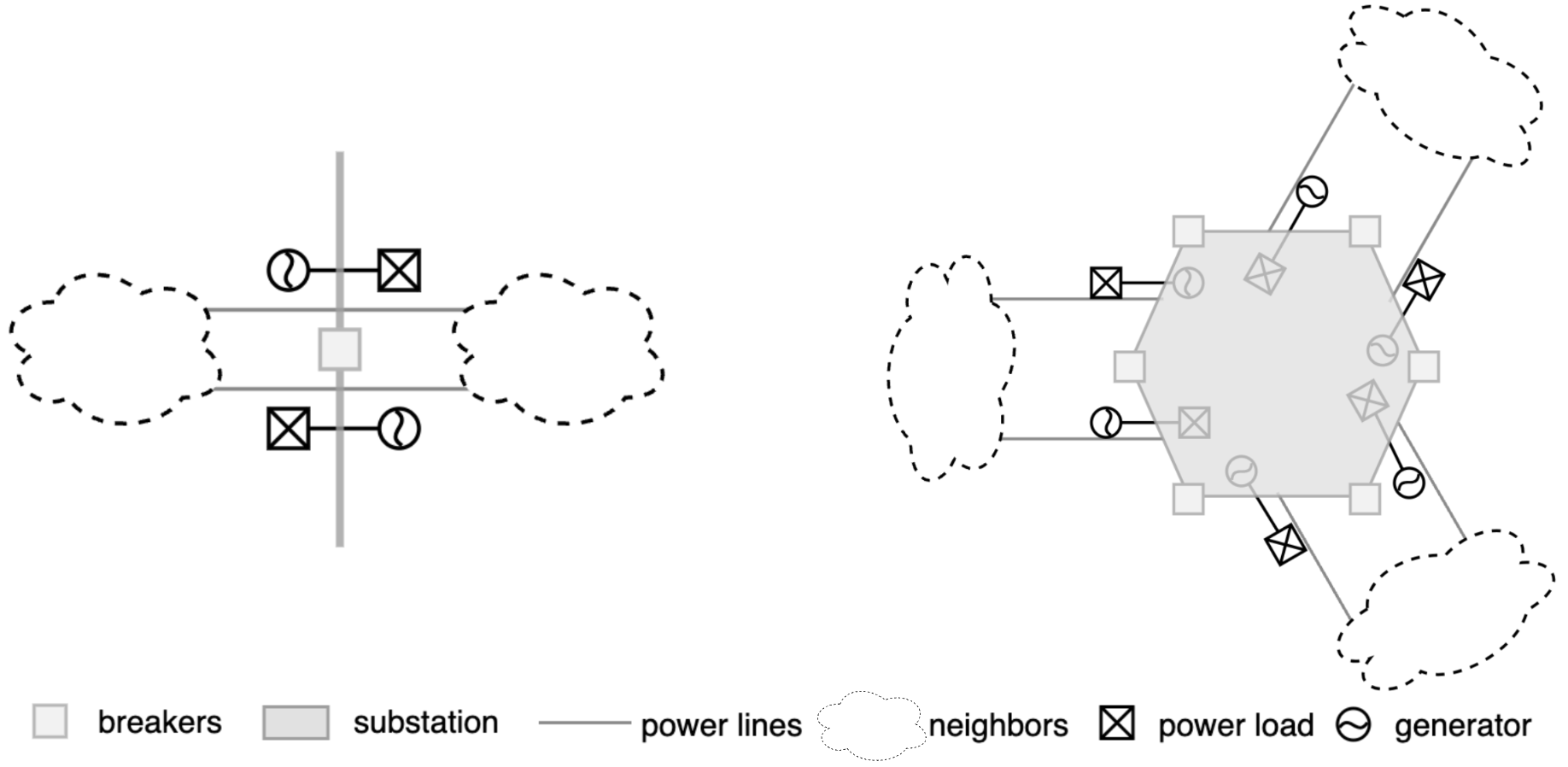}
    \caption{Node-breaker configuration of $2$-neighbor (left) and $3$-neighbor (right) substations.}
    \label{fig:neighbor_examples}
\end{figure}

To illustrate the hierarchical structure of the power grid, the paper uses diagrams showing connectivity from the system level down to individual substations. Let \( N(s) \) denote the set of neighboring substations connected to substation \( s \) (Figure~\ref{fig:neighbor_examples}). A substation with \( |N(s)| = 2 \) has a single breaker connecting two busbars, typically with one associated with generation and the other with load. A substation with \( |N(s)| \geq 3 \) follows a ring structure, consisting of \( 2|N(s)| \) breakers and \( 2|N(s)| \) busbars arranged in alternating generation and load roles.
 Busbars within a substation are connected via breakers. If a breaker between busbars \( i \) and \( j \) is open, then the voltage angles \( \theta_i \) and \( \theta_j \) are unconstrained, and no power flows between them (i.e., \( f_{ij} = 0 \)). If the breaker is closed, then \( \theta_i = \theta_j \), and the power flow \( f_{ij} \) becomes a free variable. A substation is considered split if its busbars are separated by open breakers. A substation with one breaker is split if that breaker is open. In multi-breaker substations, a split occurs when exactly two breakers are open. Following the standard substation control practices, each substation is split into at most two components, and two adjacent breakers are not allowed to be opened simultaneously to avoid isolated busbars.

\paragraph{Constraint Optimization Formulation}
\label{sec:formulation}

\begin{figure}[!t]
\begin{align}
\text{max} \quad & \lambda \tag{3} \label{eq:opt_problem} \\
\text{s.t.} \quad 
& \mu = \alpha \lambda + \beta & \tag{3a} \label{eq:scaling} \\
& P_i^G =
\begin{cases}
\lambda \cdot \bar{P}_i & \text{if } i \in \mathcal{B}_1 \\
\bar{P}_i & \text{if } i \in \mathcal{B}_2
\end{cases} & \tag{3b} \label{eq:pg} \\
& P_i^D =
\begin{cases}
d_i & \text{if } i \in \mathcal{B}_1 \\
\mu \cdot d_i & \text{if } i \in \mathcal{B}_2
\end{cases} & \tag{3c} \label{eq:pd} \\
& f_\ell = \frac{1}{X_\ell}(\theta_i - \theta_j)
\quad \forall \ell = (i,j) \in \mathcal{L} & \tag{3d} \label{eq:dcflow} \\
& -\bar{F}_\ell \le f_\ell \le \bar{F}_\ell
\quad \forall \ell \in \mathcal{L} & \tag{3e} \label{eq:flowlimit} \\
& f_r + f_{r'} = 0, \quad
|f_r| \le M \cdot z_r
\quad \forall r \in \mathcal{R} & \tag{3f} \label{eq:breakerflow} \\
& \theta_i - \theta_j \le M (1 - z_r)
\quad \forall r = (i, j) \in \mathcal{R} & \tag{3g} \label{eq:angle1} \\
& \theta_j - \theta_i \le M (1 - z_r)
\quad \forall r = (i, j) \in \mathcal{R} & \tag{3h} \label{eq:angle2} \\
& P_i^G - P_i^D + \sum_{\ell \in \mathcal{L}_i} f_\ell + \sum_{r \in \mathcal{R}_i} f_r = 0
\quad \forall i \in \mathcal{B} & \tag{3i} \label{eq:powerbalance} \\
& \theta_1 = 0 & \tag{3j} \label{eq:slack} \\
& \sum_{r \in \mathcal{R}_i} A_{ir} z_r \ge 1
\quad \forall i \in \mathcal{B} & \tag{3k} \label{eq:nodangling} \\
& \text{where } A_{ir} = 1 \text{ iff } r \text{ is connected to } i, \notag \\
& \sum_{r \in \mathcal{R}_s} z_r \ge |\mathcal{R}_s| - 2
\quad \forall s \in \mathcal{S},\ |\mathcal{R}_s| > 2 & \tag{3l} \label{eq:splitlimit}
\end{align}
\caption{The TNR Optimization Model.}
\label{fig:formulation}
\end{figure}

The formulation is shown in Figure \ref{fig:formulation}.
The objective \( \lambda \in [0,1] \) denotes the generation capacity utilization ratio in Zone~1, representing the maximum usable fraction of base generation \(\bar{\mathbf{P}}\) that satisfies all physical and topological constraints. Similarly, \( \mu \in [0,1] \)  represents the proportion of base demand \(\mathbf{d}\) that can be served under the same constraints. Constraint \eqref{eq:scaling} ensures consistency between the load and generation capacity utilization ratios. Constraints \eqref{eq:pg}–\eqref{eq:pd} define how generation and demand are fixed in zones 1 and 2. Constraints \eqref{eq:dcflow}–\eqref{eq:flowlimit} enforce DC power flow on transmission lines. Breaker flow and angle consistency are enforced through constraints \eqref{eq:breakerflow}–\eqref{eq:angle2}. Constraint \eqref{eq:powerbalance} ensures power balance at each busbar. The slack bus angle is fixed by \eqref{eq:slack}. Structural constraints \eqref{eq:nodangling} and \eqref{eq:splitlimit} ensure busbars remain connected and substations are not split into more than two components.

\section{Framework}

To address the computational complexity of the bus splitting problem, this work proposes a modular framework called \textsc{OptiGridML}, which consists of three stages: pretraining, decision learning, and inference-time repair. First, the line graph neural network (LGNN) pretraining learns physically meaningful line-flow embeddings that reflect the underlying DC power flow structure (Section~\ref{subsec:LGNN}). These flow-aware embeddings are then used to guide a heterogeneous graph neural network (HeteroGNN) that predicts substation-level breaker decisions under structural and physical constraints (Sections~\ref{subsec:heteroGNN}–\ref{subsec:integration}). Finally, an inference-time repair mechanism is applied to eliminate residual constraint violations, ensuring that the resulting topology is fully feasible and deployable (Section~\ref{subsec:repair}).

\subsection{LGNN Pretraining for Line Flows}
\label{subsec:LGNN}

\begin{figure*}[t]
  \centering
  \includegraphics[width=0.8\textwidth]{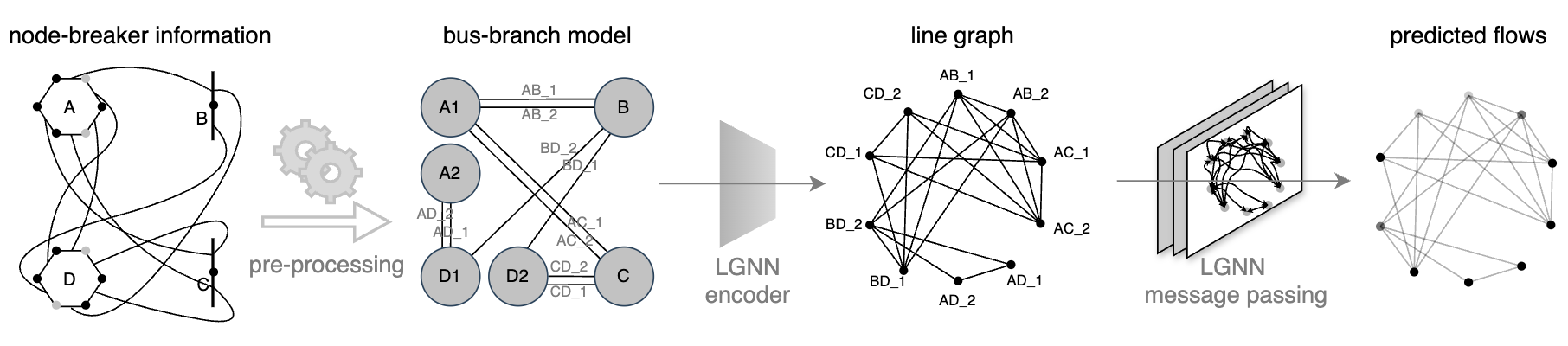}
  \caption{LGNN pretraining module that learns line-flow embeddings using power flow structure.}
  \label{fig:LGNN}
\end{figure*}

The first stage constructs and pretrains the LGNN to predict DC power flows given a fixed breaker configuration as illustrated in Figure \ref{fig:LGNN}. Supervised training is conducted on feasible solutions obtained by solving the optimization problem in Section~\ref{sec:formulation}. Each instance includes a breaker configuration \( \mathbf{z}_r \), DCOPF capacity utilization ratios \( \lambda \) and \( \mu \), and the resulting redistributed injections \( \mathbf{P}^G \) and \( \mathbf{P}^D \). 

To construct the graph representation used by the LGNN, the node-breaker model is first converted into a bus-branch model. In this abstraction, each substation is represented by a single node unless it is split, in which case it is modeled as two separate connected components. These components define the nodes \( \mathcal{B} \) of the bus-branch graph \( G = (\mathcal{B}, \mathcal{L}) \), with lines \( \mathcal{L} \) as edges. The LGNN operates on the line graph \( L(G) \), where each node corresponds to a transmission line \( \ell = (i, j) \in \mathcal{L} \), and an edge connects two lines if they share a common bus node. This representation satisfies two key requirements: it preserves fixed input dimensionality despite substation split as the number of lines is fixed, and it enables flow redistribution to be modeled directly through line-level message passing.

\subsubsection*{Line-Graph Representation}
Each transmission line \( \ell = (i, j) \in \mathcal{L} \) becomes a node in the line graph \( L(G) \). Parallel lines between the same substations are treated as distinct nodes in \( L(G) \), preserving their electrical properties.

Each line-node is initialized with feature vector $\mathbf{x}_\ell$, as:
\[
\mathbf{x}_\ell = \left[ X_\ell,\ \bar{F}_\ell,\ \sum_{k \in C_i} (P_k^G - P_k^D),\ \sum_{k \in C_j} (P_k^G - P_k^D) \right]
\]
where \( X_\ell \) is the line reactance, \( \bar{F}_\ell \) is the thermal limit, and the last two entries are aggregated net injections at the endpoint substations (connected components \( C_i \) and \( C_j \)).

An edge in \( L(G) \) connects two lines if they share at least one substation node in the bus-branch model. Let \( S_{\ell_1,\ell_2} \subseteq \mathcal{B} \) be the set of shared nodes between lines \( \ell_1 \) and \( \ell_2 \). The edge feature aggregates information over all shared substations:
\[
\mathbf{e}_{\ell_1,\ell_2} = 
\left[
\sum_{s \in S} \text{inj}(s),\;
\max_{s \in S} \text{split}(s)
\right]
\]
where 
\( \text{inj}(s) = \sum_{k \in C_s} (P_k^G - P_k^D) \), and 
\( \text{split}(s) \in \{0,1\} \) indicates whether the substation containing node \( s \) is split.
\subsubsection*{Message Passing}
Let \( h_\ell^{(t)} \in \mathbb{R}^d \) denote the embedding of node \( \ell \) at layer \( t \). The message from a neighboring line \( p \in N(\ell) \), connected via one or more shared substation nodes, is computed using:
\[
\phi_{\text{msg}}(h_\ell^{(t)}, h_p^{(t)}, \mathbf{e}_{\ell,p}) = \text{MLP}_{\text{msg}}^{(t)}\left([h_\ell^{(t)} \,\|\, h_p^{(t)} \,\|\, \mathbf{e}_{\ell,p}]\right)
\]
where \( \text{MLP}_{\text{msg}}^{(t)} \) is a 2-layer multilayer perceptron with ReLU activation, and \( \| \cdot \| \) denotes concatenation.\\
Messages are aggregated from all neighbors:
\[
m_\ell^{(t)} = \sum_{p \in N(\ell)} \phi_{\text{msg}}(h_\ell^{(t)}, h_p^{(t)}, \mathbf{e}_{\ell,p})
\]
The node embedding is updated using:
\[
h_\ell^{(t+1)} = \phi_{\text{upd}}^{(t)}\left(h_\ell^{(t)},\ m_\ell^{(t)}\right),
\]
a 2-layer MLP with ReLU activation. After \( T \) layers, each final embedding \( h_\ell^{(T)} \) is passed through a shared linear decoder to predict the flow \( \hat{f}_\ell \in \mathbb{R} \).

\subsubsection*{Flow Loss}
The LGNN is trained to minimize the mean squared error between predicted and true DC flows:
\[
L_{\text{LGNN}} = \frac{1}{|\mathcal{L}|} \sum_{\ell \in \mathcal{L}} \left( \hat{f}_\ell - f_\ell \right)^2
\]
where \( f_\ell = f_\ell^{\text{DCOPF}} \) is the ground-truth flow computed from the DCOPF solution. Once trained, the LGNN defines a flow encoder function \( \psi \) that maps a given network topology and power injection profile to approximate line flows:
\[
\psi : (\mathbf{z}_r,\, \bar{\mathbf{P}},\, \mathbf{d}) \mapsto \hat{\mathbf{f}} \in \mathbb{R}^{|\mathcal{L}|}
\]
This encoder is later integrated into the HeteroGNN framework (Section~\ref{subsec:integration}) to provide flow-aware supervision for breaker predictions.

\subsection{HeteroGNN for Optimal Breaker Status}
\label{subsec:heteroGNN}

\begin{figure}[t]
  \centering
  \includegraphics[width=0.47\textwidth]{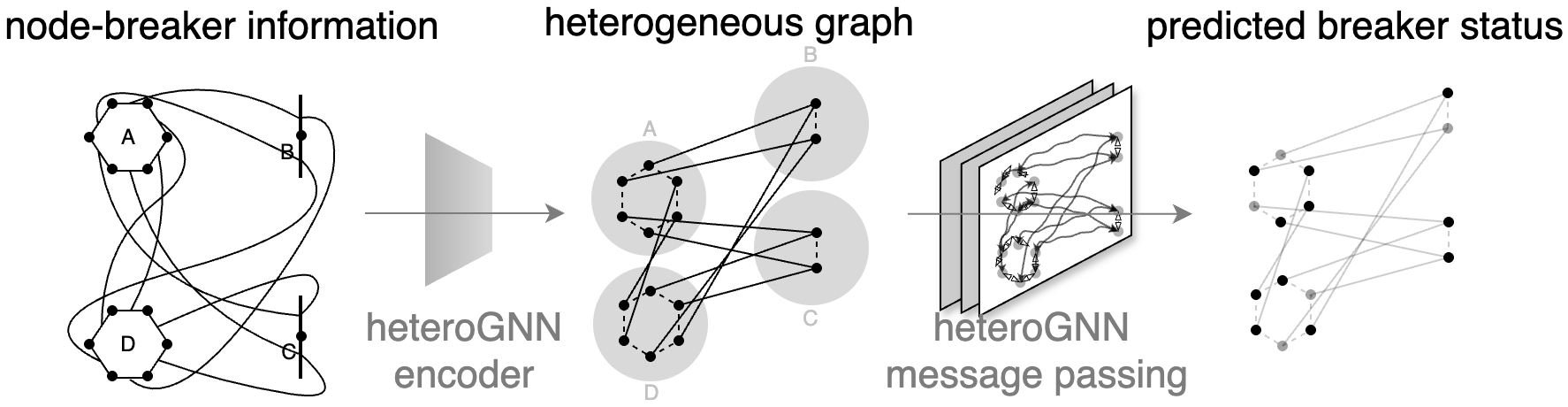}
  \caption{Heterogeneous GNN that predicts breaker decisions \( \hat{z}_r \in (0,1) \) for substation edges.}
  \label{fig:heteroGNN}
\end{figure}

This component predicts the binary breaker status \( z_r \in \{0,1\} \) for each internal substation connection, given the original, unsplit grid information. The goal is to approximate the optimal configuration derived from solving the optimization in Section~\ref{sec:formulation}. Unlike classical GNNs that treat all edges homogeneously, this setting requires differentiating internal breaker connections from external transmission lines and learning edge-level predictions. A type-aware GNN is constructed as illustrated in Figure~\ref{fig:heteroGNN} to produce breaker decisions that conform to physical and topological constraints.

% \begin{figure*}[t]
%   \centering
%   \includegraphics[width=0.7\textwidth]{figures/framework/fig_heteroGNN.png}
%   \caption{Heterogeneous GNN that predicts breaker decisions \( \hat{z}_r \in (0,1) \) for substation edges using bus-breaker topology.}
%   \label{fig:heteroGNN}
% \end{figure*}

\subsubsection*{Bus-Breaker Graph Encoding}

The power grid is modeled as a global heterogeneous graph \( \mathcal{G} = (\mathcal{V}, \mathcal{E}) \), where each node \( i \in \mathcal{V} \) corresponds to a busbar. There is one node type (busbar) and two edge types:
\begin{itemize}
    \item \textit{Breaker edges} \( (i,j) \in \mathcal{E}_{\text{br}} \) represent breaker \( r \in \mathcal{R} \) connecting busbars within a substation.
    \item \textit{Transmission line edges} \( (i,j) \in \mathcal{E}_{\text{line}} \) represent the physical lines \( \ell \in \mathcal{L} \) connecting different substations.
\end{itemize}

\noindent\textbf{Node features:}
Each busbar node \( i \in \mathcal{B} \) is initialized with:
\[
\mathbf{x}_i = \left[ P_i^G,\; P_i^D,\; \text{deg}^{\text{line}}_i,\; \text{deg}^{\text{total}}_i \right] \in \mathbb{R}^4
\]
where \( \text{deg}^{\text{line}}_i \) is the number of transmission lines incident to \( i \), and \( \text{deg}^{\text{total}}_i \) is the number of incident edges (transmission lines and breakers).

\noindent\textbf{Edge features:} A breaker edge \( {(i,j)}\) is initialized with:
    \[
    \mathbf{e}_{ij}^{\text{br}} = \left[ z_r,\; \text{conn}_L,\; \text{conn}_R \right] \in \mathbb{R}^3
    \]
    where \( z_r \in \{0,1\} \) is the breaker status, and \( \text{conn}_L \), \( \text{conn}_R \) denote the number of \emph{active connections} - closed neighboring breakers and connected transmission lines, at busbars \( i \) and \( j \), respectively, excluding \( z_r \) itself.

\noindent A line edge \( (i,j) \in \mathcal{E}_{\text{line}} \) is initialized with:
    \[
    \mathbf{e}_{ij}^{\text{line}} = \left[ X_\ell,\; \bar{F}_\ell \right] \in \mathbb{R}^2
    \]
    where \( X_\ell \) is the line reactance and \( \bar{F}_\ell \) is its thermal limit.

\subsubsection*{Message Passing in HeteroGNN}

Separate message functions are used for each edge type. Let \( h_i^{(t)} \in \mathbb{R}^d \) be the embedding of node \( i \) at layer \( t \). The type-specific aggregated messages are:
\[
M_i^{\text{br}} = \sum_{j \in \mathcal{N}_i^{\text{br}}} \phi^{\text{br}}(h_i^{(t)}, h_j^{(t)}, e_{ij}^{\text{br}})
\]
\[
M_i^{\text{line}} = \sum_{j \in \mathcal{N}_i^{\text{line}}} \phi^{\text{line}}(h_i^{(t)}, h_j^{(t)}, e_{ij}^{\text{line}})
\]

The combined message is:
\[
M_i = \alpha_{\text{br}} \cdot M_i^{\text{br}} + \alpha_{\text{line}} \cdot M_i^{\text{line}}, \quad
\alpha_k = \frac{\exp(w_k^\top \bar{M}_i^k)}{\sum_{l} \exp(w_l^\top \bar{M}_i^l)}
\]
where \( \bar{M}_i^k \) is a pooled summary of messages from edge type \( k \), and \( w_k \) are learned attention weights.

Each node’s embedding is updated via a gated recurrent unit (GRU):
\[
h_i^{(t+1)} = \text{GRU}(M_i,\, h_i^{(t)})
\]

\subsubsection*{Breaker Prediction}

After \( T \) rounds of message passing, each breaker edge \( r = (i, j) \in \mathcal{E}_{\text{br}} \) is assigned a prediction:
\[
\hat{z}_r = \sigma\left( \text{MLP}_{\text{pred}} \left([h_i \,\|\, h_j \,\|\, e_{ij}^{\text{br}}] \right) \right)
\]
where \( \text{MLP}_{\text{pred}} \) is a shared two-layer MLP with ReLU activations, and \( \sigma \) is the sigmoid function. Predictions are made for all breaker edges.

\subsubsection*{Breaker Prediction Loss}
Let \( z_r \in \{0,1\} \) be the ground truth status and \( \hat{z}_r \in (0,1) \) the predicted probability. The binary cross-entropy loss is:
\[
L_{\text{br}} = \frac{1}{|\mathcal{E}_{\text{br}}|} \sum_{r \in \mathcal{E}_{\text{br}}}
\left[ -z_r \log(\hat{z}_r) - (1 - z_r) \log(1 - \hat{z}_r) \right]
\]

\subsection{Integration of the Flow Encoder}
\label{subsec:integration}

The pretrained LGNN flow encoder is integrated into the HeteroGNN framework to impose physical structure on predicted breaker configurations. Although line flow labels are unavailable for arbitrary predicted topologies, the LGNN provides approximate DC flow estimates that support physics-informed supervision.

\subsubsection*{Flow-Aware Auxiliary Supervision}

To evaluate a predicted breaker configuration \( \hat{\mathbf{z}}_r \), the induced bus-branch topology is constructed by identifying connected components within each substation. Rather than solving a new DCOPF for each predicted topology, the same capacity utilization ratios \( \lambda \) and \( \mu \) from the labeled optimal topology are reused for auxiliary loss. For each substation \( S \), with base generation \( \bar{P}_S \) and demand \( d_S \), the scaled generation and load are redistributed evenly across each busbar \( i \in C \subseteq \mathcal{B}_S \) in its connected components:
\[
\hat{P}^G_i = \frac{1}{|C|} \cdot \lambda \cdot \bar{P}_S, \qquad
\hat{P}^D_i = \frac{1}{|C|} \cdot \mu \cdot d_S
\]
These injections and the predicted topology are passed to the pretrained flow encoder \( \psi \), yielding approximate line flows:
\[
\hat{\mathbf{f}} = \psi(\hat{\mathbf{z}}_r,\, \bar{\mathbf{P}},\, \mathbf{d})
\]

\subsubsection*{Flow Consistency Loss}

Because the true DC flows corresponding to \( \hat{\mathbf{z}}_r \) are unknown, a flow consistency loss is introduced for Kirchhoff’s Current Law. The loss penalizes mismatch between predicted flows and nodal net injections:
\[
L_{\text{flow}} = \sum_{i \in \mathcal{B}} \left( \sum_{j \in \mathcal{N}(i)} \hat{f}_{ij} - (\hat{P}^G_i - \hat{P}^D_i) \right)^2
\]
This loss approximates $L_{\text{LGNN}}$ and penalizes predicted breaker configurations that result in redispatched flows violating nodal balance constraints.

\subsection{Training Loss with Constraint Satisfaction}
\label{subsec:training_loss}

The HeteroGNN is trained to predict breaker configurations using a multi-objective loss that combines classification accuracy, physical consistency, and constraint satisfaction.

\subsubsection*{Feasibility Penalties}

To enforce physical and topological feasibility, we define auxiliary penalties:

\begin{itemize}
    \item Thermal limit violation:
    \[
    L_{\text{cap}} = \frac{1}{|\mathcal{L}|} \sum_{\ell \in \mathcal{L}} \left[
    \max\left(0, \frac{|\hat{f}_\ell| - \bar{F}_\ell}{\bar{F}_\ell} \right)
    \right]^2
    \]
    
    \item Dangling busbars:
    \[
    L_{\text{dangling}} = \sum_{i \in \mathcal{B}} \mathds{1}\left( \deg^{\text{br}}_i = 0 \right)
    \]
    
    \item Excessive substation splits:
    \[
    L_{\text{group}} = \sum_{s \in \mathcal{S}} \max\left(0,\ \text{Groups}(s;\hat{\mathbf{z}}_r) - 2 \right)
    \]
    where \text{Groups}$(s; \hat{\mathbf{z}}_r)$ returns the number of connected components within substation $s$.
\end{itemize}
The total feasibility penalty is:
\[
L_{\text{feasibility}} = L_{\text{cap}} + L_{\text{dangling}} + L_{\text{group}}
\]

\subsubsection*{Total Loss Function}

The final loss combines breaker prediction with auxiliary supervision:
\[
L = L_{\text{br}} + \rho_1 L_{\text{flow}} + \rho_2 L_{\text{feasibility}}
\]
where \( \rho_1, \rho_2 \) are weights controlling the influence of flow regularization and feasibility constraints.

\begin{table*}[t]
\centering
\small
\begin{tabular}{|c|c|c|c|c|c|c|}
\hline
\textbf{Breakers} & \textbf{Substations} & \textbf{\#Neighbors/Substation} & \textbf{\#Samples} & \textbf{Method} & \textbf{Avg. Runtime} & \textbf{Std. Dev.} \\
\hline
50   & 12   & 2 - 4 & 500   & MIP               & 27.9s  & 37.42s \\
100  & 20   & 2 - 5 & 1,000 & MIP               & 11.2m  & 3.42m \\
500  & 100  & 2 - 6 & 2,000 & MIP               & 10.7h  & 2.44h \\
1000 & 200  & 2 - 7 & 5,000 & Heuristic approx. & 5.9h   & 0.01h \\
\hline
\end{tabular}
\caption{Summary of generated networks and instances, with runtime per instance.}
\label{tab: data}
\end{table*}

\begin{table*}[t]
\centering
\small
\begin{tabular}{|c|cc|cc|cc|cc|cc|}
\hline
\multirow{3}{*}{Breakers} &
\multicolumn{4}{c|}{\textbf{ML Baselines}} &
\multicolumn{6}{c|}{\textbf{\textsc{OptiGridML} Variants} (our method)} \\
\cline{2-11}
& \multicolumn{2}{c|}{MLP} &
  \multicolumn{2}{c|}{GCN} &
  \multicolumn{2}{c|}{\(\rho_1{=}0\) (no flow loss)} &
  \multicolumn{2}{c|}{\(\rho_2{=}0\) (no feasibility)} &
  \multicolumn{2}{c|}{\(\bm{\rho_1{=}1.2,\ \rho_2{=}2.0}\)} \\
\cline{2-11}
& $\lambda$ & Viol. \% & $\lambda$ & Viol. \% & $\lambda$ &  Viol. \% & $\lambda$ &  Viol. \% & $\lambda$ & Viol. \% \\
\hline
50   & +3.1\% & 40.0\% & +6.7\% & 26.0\% & +8.4\% & 4.0\% & +13.1\% & 20.0\% & \textbf{+18.7}\% & \textbf{4.0}\% \\
100  & +3.9\% & 37.0\% & +7.3\% & 27.0\% & +9.9\% & \textbf{4.0}\% & +13.8\% & 21.0\% & \textbf{+17.9}\% & 5.0\% \\
500  & +4.6\% & 39.5\% & +8.1\% & 28.5\% & +9.5\% & 4.5\% & +14.9\% & 27.5\% & \textbf{+18.3}\% & \textbf{4.5}\% \\
1000 & +4.3\% & 40.2\% & +7.9\% & 30.4\% & +8.7\% & \textbf{5.4}\% & +12.1\% & 29.6\% & \textbf{+17.6}\% & 5.6\% \\
\hline
\end{tabular}
\caption{Breaker prediction performance. Percentage improvement $\lambda$ is reported relative to the base topology (all breakers closed).
Violation \% indicates the fraction of instances with any constraint violation under the predicted topology \textit{before} repair.}
\label{tab:heterognn_accuracy}
\end{table*}

\subsection{Inference Constraint Satisfaction and Repair}
\label{subsec:repair}

While feasibility penalties are applied during training, the HeteroGNN may predict a breaker configuration \( \hat{\mathbf{z}}_r \in \{0,1\}^k \) that violates hard constraints at inference time. To ensure that the output is operationally valid under the formulation in Section~\ref{sec:formulation}, a lightweight post-processing repair step is introduced. Let \( \hat{\mathcal{G}} \) denote the bus-breaker graph induced by \( \hat{\mathbf{z}}_r \). The following feasibility checks are performed:
\begin{itemize}
    \item \textit{Dangling busbars:} For any busbar \( i \in \mathcal{B} \) with zero incident closed breakers, i.e., \( \deg^{\text{br}}_i(\hat{\mathbf{z}}_r) = 0 \), connectivity is restored by reclosing at least one adjacent breaker.
    
    \item \textit{Excessive substation splits:} For substation \( s \in \mathcal{S} \), let \( \mathcal{B}_s \) denote its busbars and \( \mathcal{R}_s \subseteq \mathcal{E}_{\text{br}} \) its internal breakers. If the number of connected components induced by closed breakers satisfies \( |\mathcal{C}_s(\hat{\mathbf{z}}_r)| > 2 \), a minimal subset \( \mathcal{R}'_s \subseteq \mathcal{R}_s \) is reclosed to enforce \( |\mathcal{C}_s(\hat{\mathbf{z}}_r + \mathds{1}_{\mathcal{R}'_s})| = 2 \).
    
    \item \textit{Thermal flow violations:} Let \( \hat{\mathbf{f}} \) be the flows estimated from the LGNN under \( \hat{\mathbf{z}}_r \); define the overloaded lines as
    \[
    \mathcal{L}_{\text{viol}} = \left\{ \ell \in \mathcal{L} : |\hat{f}_\ell| > \bar{F}_\ell \right\}.
    \]
    For each \( \ell = (i,j) \in \mathcal{L}_{\text{viol}} \), nearby internal breakers in the substations containing \( i \) or \( j \) are adjusted to reduce congestion, prioritizing configurations that minimally impact overall power export.
\end{itemize}

\noindent Since each substation has at most \( O(|\mathcal{B}_s|^2) \) possible internal connections and violations are local, the repair process terminates in polynomial time. The resulting configuration \( \hat{\mathbf{z}}_r^{\text{feas}} \) satisfies all structural and physical constraints.

\subsubsection*{Post-Repair DCOPF Evaluation}

A DCOPF is solved under the repaired topology \( \hat{\mathbf{z}}_r^{\text{feas}} \), yielding a physically consistent estimate of the capacity utilization ratio \( \lambda \), which serves as the final performance metric and quantifies the maximum achievable power transfer under the predicted configuration.

\section{Experiments}

\paragraph{Dataset Description and Parameter Settings}

The dataset consists of synthetic power networks of varying sizes, up to 1000 breakers and 200 substations. Each substation has 2 to 7 neighboring substations connected via double power lines. Each network is labeled with optimal or near optimal breaker configurations for maximizing power export under DC power flow constraints (Table~\ref{tab: data}). For each network topology, multiple input instances are generated by randomly perturbing the generation and load values of each substation by $\pm20\%$ and rebalancing the system. Ground-truth breaker labels are generated for small networks with up to 500 breakers using an exact MIP solver, Gurobi, a commercial optimization software. For larger networks where the MIP solver did not find a solution, a faster heuristic search method is used to approximate near-optimal breaker configurations followed by feasibility validation via DC-OPF. The heuristic approximation was evaluated for accuracy in small networks by comparing with MIP solver. On average, the heuristic approximation could return feasible topology with $96.0\%$ optimal $\lambda$ for 500-breaker networks in $48$ minutes per instance. The generated dataset was then divided into training, validation,
and testing subsets in an 8:1:1 ratio. Table~\ref{tab:params} summarizes the key hyperparameters and architecture choices used for the LGNN flow encoder pretraining and the HeteroGNN breaker predictor training, determined through hyperparameter tuning.

\begin{table}[t]
\centering
\small
\begin{tabular}{lcc}
\toprule
\textbf{Hyperparameter} & \textbf{LGNN} & \textbf{HeteroGNN} \\
\midrule
Message Passing Layers     & 3                 & 4 \\
Hidden Dimension & 64              & 64 \\
Learning Rate              & Adam $1 \times 10^{-3}$ & Adam $1 \times 10^{-3}$ \\
Batch Size        & 32  & 16\\
Epochs          & 200 (early stop) & 300 (early stop)\\ 
Loss Weights $\rho_1$, $\rho_2$ & -           & 1.2, 2.0 \\
\bottomrule
\end{tabular}
\caption{Architecture and training hyperparameters.}
\label{tab:params}
\end{table}

\paragraph{Prediction Accuracy}

The experiments evaluate \textsc{OptiGridML}’s ability to recover near-optimal breaker configurations (Table~\ref{tab:heterognn_accuracy}). Vanilla ML models are used as baselines. 
The \textbf{MLP baseline} is a fully connected network operating on flattened features without graph structure. Without any topological information, about 40\% tested instances violated one or more structural constraints \textit{before} repairing using rules in Section~\ref{subsec:repair}. Once repaired and used as an input into a DCOPF solver, the topology predicted by the MLP baseline returned capacity utilization ratio $\lambda$ only about 4\% higher than $\lambda$ in the original topology without breaker openings. 
 The \textbf{GCN baseline} is constructed without heterogeneous edge types or substation-specific constraints. It captures more structural information than the MLP model, but still fails to predict feasible topologies for about 27\% instances \textit{before} repairing.
Ablation studies performed on \textsc{OptiGridML} demonstrate the impact of flow awareness and constraint satisfaction. By setting \(\rho_1 = 0\), flow loss is ignored and only feasibility is considered. As a result, only 4\% instances need to go through the repair process, but $\lambda$ is not close to optimal. Setting \(\rho_2 = 0\) and considering only flow embeddings causes more feasibility violations, which leads to higher $\lambda$ after resolving the violations. By setting $\rho_1=1.2$, $\rho_2=2.0$, \textsc{OptiGridML} successfully predicts feasible breaker configurations for 95\% instances, and the rest goes through a fast repair (Section~\ref{subsec:repair}). As a result, \textsc{OptiGridML}'s predicted and repaired configurations raised $\lambda$ by 18\% from the base topology. Instance-based results are presented in the appendix. 

\paragraph{Computational Efficiency}

Table~\ref{tab:gnn_vs_mip} summarizes the training cost and average inference time of the HeteroGNN with a pretrained LGNN across varying network sizes. Once trained, \textsc{OptiGridML} predicts breaker configurations over 10{,}000 times faster than solving a MIP for a 100-breaker network, and over 2 million times faster for a 500-breaker network. The predicted capacity utilization ratio \( \hat{\lambda} \) reaches over 90\% of the optimal \( \lambda^* \) obtained from MIP. On larger networks with 1{,}000 breakers---where MIP fails to return a solution within the time limit---\textsc{OptiGridML} produces a feasible, high-quality topology in just 228 milliseconds.

\begin{table}[t]
\centering
\small
\begin{tabular}{|c|c|c|c|c|}
\hline
\textbf{Breakers} & \textbf{Train} &\shortstack{\textbf{Inference} \\ \textbf{per inst}}  & \shortstack{\textbf{Speedup} \\ \textbf{vs. MIP }} & \shortstack{\textbf{Accuracy} \\ \textbf{vs. MIP $\bm{\lambda^*}$}}  \\
\hline
50   & 3 min  & 43 ms  & 649$\times$  & 91.2\% \\
100  & 7 min  & 67 ms  & 10,030$\times$ & 90.8\% \\
500  & 33 min & 153 ms & 251,765$\times$ & 88.3\% \\
1000 & 62 min & 228 ms & - & - \\
\hline
\end{tabular}
\caption{Comparison of HeteroGNN vs MIP performance. MIP is intractable for instances with over 500 breakers.}
\label{tab:gnn_vs_mip}
\end{table}

\section{Conclusions and Discussions}
% This work introduced OptiGridML, a novel machine learning framework designed for the discrete topology optimization problem in transmission network reconfiguration (TNR). By combining a pretrained line-graph neural network (LGNN) for DC power flow approximation and a heterogeneous GNN (HeteroGNN) for breaker configuration prediction, OptiGridML demonstrated the feasibility of fast and accurate topology decisions, significantly outperforming traditional optimization methods.

% Experiments conducted on synthetic networks comprising up to 1,000 breakers validated the effectiveness of the proposed approach. OptiGridML achieved substantial improvements, enhancing the export capacity by up to 18\% compared to baseline topologies, while dramatically reducing inference times from hours (using conventional mixed-integer programming methods) to milliseconds. These results underscore the practical potential of leveraging physics-informed and structured machine learning techniques for operational decision-making in power grids.

% The integration of a physics-informed consistency loss and constraint-aware training strategy played a critical role in ensuring both accuracy and feasibility. The lightweight rule-based repair mechanism further reinforced operational validity, guaranteeing feasibility with negligible computational overhead.

This paper presented \textsc{OptiGridML}, a novel machine learning framework for substation-level transmission network reconfiguration that replaces expensive MIP solves with structured GNN inference. The framework combines a pretrained line-graph flow encoder (LGNN) with a heterogeneous GNN (HeteroGNN) to predict breaker-level topology reconfigurations, guided by a loss that enforces Kirchhoff’s laws without requiring ground-truth flows. Experimental results on synthetic transmission networks (inspired by real-life settings) with up to 1,000 breakers demonstrate that \textsc{OptiGridML} produce feasible configurations that increase power export by up to 18\% over the base topology, while reducing inference time from hours to milliseconds.

Several directions remain open for future research. Extending OptiGridML to real-world power system dataset could further validate its practical applicability. Exploring real-time settings under varying load and generation conditions represents a promising next step. Lastly, investigating more complex constraint handling through advanced differentiable optimization layers or reinforcement learning-based repair strategies may improve operational effectiveness.

\bibliography{aaai2026}

\end{document}